\documentclass[journal]{IEEEtran}
\usepackage{xcolor}
\usepackage{setspace}
\usepackage{float}
\usepackage[fleqn]{amsmath}
\usepackage{enumerate}%\begin{enumerate}[i] % reacts to 1, a, A, i, and I
\usepackage{mathrsfs}
\usepackage{makeidx}         % allows index generation
\usepackage{graphicx}        % standard LaTeX graphics tool
\usepackage{multicol}        % used for the two-column index
\usepackage{subfigure}
\usepackage{epsfig,amssymb,latexsym}
\usepackage{psfrag}
\usepackage[ruled]{algorithm}
\usepackage{algorithmic}
\usepackage{url}
\usepackage{amsmath}
\usepackage{amssymb}

\newtheorem{remark}{Remark}%[section]
\newcommand{\be}{\begin{equation}}
\newcommand{\ee}{\end{equation}}
\newcommand{\bea}{\begin{eqnarray}}
\newcommand{\eea}{\end{eqnarray}}
\newcommand{\bal}{\begin{align}}
\newcommand{\eal}{\end{align}}
\newcommand{\ba}{\begin{array}}
\newcommand{\ea}{\end{array}}
\newcommand{\bc}{\begin{center}}
\newcommand{\ec}{\end{center}}

\newcommand{\argmax}{\arg\!\max}
\allowdisplaybreaks
\ifCLASSINFOpdf
\else
\fi
\begin{document}
%
% paper title
% Titles are generally capitalized except for words such as a, an, and, as,
% at, but, by, for, in, nor, of, on, or, the, to and up, which are usually
% not capitalized unless they are the first or last word of the title.
% Linebreaks \\ can be used within to get better formatting as desired.
% Do not put math or special symbols in the title.
\title{Saliency Guided Hierarchical Robust Visual Tracking}
%
%
% author names and IEEE memberships
% note positions of commas and nonbreaking spaces ( ~ ) LaTeX will not break
% a structure at a ~ so this keeps an author's name from being broken across
% two lines.
% use \thanks{} to gain access to the first footnote area
% a separate \thanks must be used for each paragraph as LaTeX2e's \thanks
% was not built to handle multiple paragraphs
%

\author{Fangwen~Tu,
        Shuzhi~Sam~Ge,~\IEEEmembership{Fellow,~IEEE,}
        Yazhe~Tang,
        and~Chang~Chieh~Hang,~\IEEEmembership{Fellow,~IEEE}% <-this % stops a space
%\thanks{F. Tu is with the Department of Electrical and Computer Engineering, National
%University of Singapore, Singapore 117576 (e-mail: $\text{fangwen\_tu@u.nus.edu}$).}% <-this % stops a space
%\thanks{S. S. Ge is with the Department of Electrical and Computer Engineering, National
%University of Singapore, Singapore 117576 and also with the Social Robotics Lab, Interactive Digital
%Media Institute (IDMI), National University of Singapore, Singapore 117576, (e-mail: samge@nus.edu.sg).}% <-this % stops a space
%\thanks{Y. Tang is with the Department of Precision Mechanical Engineering, Shanghai University, Shanghai. He is also with the Temasek Laboratories, National University of Singapore, Singapore (yztang2008@yahoo.com).}% <-this % stops a space
%\thanks{C. C. Hang is with the Department of Electrical and Computer Engineering, National
%University of Singapore, Singapore 117576 (e-mail: etmhead@nus.edu.sg).}
}
\maketitle

% As a general rule, do not put math, special symbols or citations
% in the abstract or keywords.
\begin{abstract}
A saliency guided hierarchical visual tracking (SHT) algorithm containing global and local search phases is proposed in this paper. In global search, a top-down saliency model is novelly developed to handle abrupt motion and appearance variation problems. Nineteen feature maps are extracted first and combined with online learnt weights to produce the final saliency map and estimated target locations. After the evaluation of integration mechanism, the optimum candidate patch is passed to the local search. In local search, a superpixel based HSV histogram matching is performed jointly with an L2-RLS tracker to take both color distribution and holistic appearance feature of the object into consideration. Furthermore, a linear refinement search process with fast iterative solver is implemented to attenuate the possible negative influence of dominant particles. Both qualitative and quantitative experiments are conducted on a series of challenging image sequences. The superior performance of the proposed method over other state-of-the-art algorithms is demonstrated by comparative study.\\
\end{abstract}

% Note that keywords are not normally used for peerreview papers.
\begin{IEEEkeywords}
Top-down saliency model, superpixel histogram matching, linear refinement search, hierarchical structure, visual tracking
\end{IEEEkeywords}

% For peer review papers, you can put extra information on the cover
% page as needed:
% \ifCLASSOPTIONpeerreview
% \begin{center} \bfseries EDICS Category: 3-BBND \end{center}
% \fi
%
% For peerreview papers, this IEEEtran command inserts a page break and
% creates the second title. It will be ignored for other modes.
\IEEEpeerreviewmaketitle

\section{Introduction}
Visual tracking aiming at estimating the location of specified target in video sequences is one of the most important research branch in computer vision. It has wide applications including security surveillance, robotics, motion analysis, military patrol and etc. Although, much attention has been attracted to this topic and there exists many breakthrough during the past few years, it is still very challenging to develop a robust algorithm because of the scene variation such as  partial or full occlusion, abrupt motion, background clutter, illumination variation and non-rigid deformation \cite{lu2013locally}.\\
\indent A popular trend to tackle visual tracking problem is to involve sparse representation since it was firstly proposed in \cite{mei2009robust}. The motion and observation model is employed and embedded into a particle filter framework with updated appearance dictionary. Their following works further enhance the speed of an L1 tracker \cite{mei2011minimum} \cite{bao2012real} by reducing the number of samples to be decomposed and proposing an accelerated proximal gradient (APG) solver. To consider the holistic and part-based information simultaneously, in \cite{zhong2014robust} a sparse discriminative classifer (SDC) and a sparse generative model (SGM) are developed jointly for visual tracking. It improves the discriminative power of the tracker at the expense of more time consumption. To alleviate the computational burden, a fast tracking algorithm using non-adaptive random projected appearance model is proposed in \cite{zhang2014fast}. The superior speed benefits a lot from the its coarse-to-fine framework. To ensure the sparsity of the coefficient, most of the state-of-the-art algorithms apply L1 regularization constraint. A L2-RLS tracker which shows competitive performance is proposed in \cite{xiao2014l2}. Compared with traditional L1 tracker, L2-RLS achieves optimizing computational complexity by only recalculating a project matrix after the template updating in solving the cost function. However, all of these algorithms only depend on the greyscale of the image and abandon the color cues of the target. In addition, their work usually assume that the target location in current frame is near from that in the last frame. Therefore, it is very difficult to handle the cases with abrupt motion.\\
\indent In this paper, we perform a saliency guided global search ahead the random walking sampling in particle filter to provide a rough location of the target. Saliency with bottom-up structure is originally used to predict the eye movement in a scene using the low-level cues (e.g., oriented filter responses and color) \cite{yang2012top}. While, to make it suitable for visual tracking in which we hope to emphasize on certain target, a top-down structure with prior information needs to be developed. Previous studies like \cite{li2016top} \cite{su2014abrupt} have verified the feasibility to incorporate top-down saliency into visual tracking. In \cite{li2016top}, the saliency map is built using a modified frequency-tuned method with pre-calculated weights in the same way of VOCUS \cite{frintrop2006vocus}. Then the target is tracked with a local and global search processes. In \cite{su2014abrupt}, the saliency map is generated by assigning weights to three conspicuity maps in Itti and Koch's saliency model \cite{itti2000saliency}. However, in their works, the weights are computed in the first frame and remain unchanged during the tracking, which makes the algorithm fail to adapt to the appearance variation of the target. Moreover, the prior information from the last frame also plays a crucial role in guiding the location prediction. Thus, to overcome these problems, a novel weight updating mechanism as well as a comprehensive saliency map generation method are proposed in this paper.\\
\indent To further take advantage of the color information, a superpixel based HSV histogram matching is newly introduced in this paper. Superpixels can be utilized to reduce the sample set that we need to consider because each superpixel patch contains pixels with similar color feature that can be merged together. Thus, it has been widely employed to address visual tracking task \cite{li2015visual} \cite{oron2015locally} \cite{wang2015visual}. In \cite{yang2014robust}, a superpixel based discriminative appearance model is constructed to facilitate the tracker to distinguish the target from the background. Although the result is promising, it misses the cue of structural information of the object which can be addressed by conventional template matching approaches. To cope with this, a joint local search scheme is proposed by combining a simplified superpixel matching with L2-RLS tracker. Superpixel matching equips the algorithm with the capability to identify the target through color distribution. Meanwhile, L2-RLS tracker will investigate the candidate patches from holistic appearance perspective.\\
\indent Finally, the hierarchical structure is completed with a linear refinement search. This search is developed to balance the contribution of particles with high confidence and avoid the domination of individual ones caused by the traditional maximum a posterior (MAP) operation. The idea is inspired by \cite{wang2010locality} and \cite{wang2015visual} which hope to extend the state space of particle observations from discrete to continuous by conducting local linear coding. Different from them, the proposed method is only applied to the selected promising candidates and hence, a customized optimization function is required. In addition, a fast iterative solver with analytical solution is established and the improved performance is demonstrated by experimental study.\\
\indent The main contributions of this work is three-fold and can be summarized as follows:
\begin{itemize}
  \item  A novel saliency guided global search algorithm is proposed considering prior information and adaptivity to object appearance variation. Nineteen features containing comprehensive description of the target are combined with online learnt weights to produce a top-down saliency map for candidate patch selection. A corresponding integration mechanism is developed to filter out the false targets. The global search can provide a rough target location prediction for the tracker to handle the abrupt motion and appearance variation problems.
  \item Superpixel based HSV histogram matching is incorporated into an L2-RLS tracker to involve the color distribution of the object and achieve a joint observation likelihood. This method can not only boost the discriminative power of an intensity template based tracker by introducing additional color cue but consider the structural information by using superpixels.
  \item Linear refinement search is designed before the final estimated result is obtained to further rectify the bounding box and lower the drifting risk of a single dominate particle caused by the MAP operation by sharing the risk with several most promising candidate patches. A novel cost function is proposed and a fast iterative solver with analytical solution in each loop is developed. The capability of this search to upgrade the accuracy is validated by case study.
\end{itemize}

\indent The organization of this paper is as follows. Section \uppercase\expandafter{\romannumeral2} briefly introduces the preliminary knowledge and concept on particle filter.
Section \uppercase\expandafter{\romannumeral3} presents the details of saliency guided global search. The integration mechanism between global search and local search is elaborated in Section \uppercase\expandafter{\romannumeral4}. Section \uppercase\expandafter{\romannumeral5} states the hierarchical local search containing superpixel matching and refinement search. Both qualitative and quantitative experiments are conducted in Section \uppercase\expandafter{\romannumeral6}. Finally, Section \uppercase\expandafter{\romannumeral7} gives some concluding remarks.
\section{Preliminaries on Particle Filter}
Bayesian inference framework is usually applied for visual tracking problem. It is supposed to estimate the posterior distribution $p(x_t|Y_t)$ of state variables characterizing a dynamic system by
\begin{eqnarray}
p(x_t|Y_t)\propto p(y_t|x_t) \int p(x_t|x_{t-1})p(x_{t-1}|Y_{t-1})\text{d}x_{t-1}
\end{eqnarray}
where $Y_t=[y_1,y_2,...,y_t]$ denotes the observation vector up to $\emph{t}$th frame, $x_t$ is the state variable of a target in frame \emph{t} and $p(x_t|x_{t-1})$ represents the motion model predicting the state in current frame with the immediate previous state, $p(y_t|x_t)$ indicates the observation model which is a likelihood function in essence. Conventionally, the estimated state $\hat{x}_t$ can be obtained by MAP operation
\begin{eqnarray}
\hat{x}_t=\arg \max_{x_t^i} p(y_t|x_t^i) p(x_t^{i}|x_{t-1})
\end{eqnarray}
where $x_t^i$ indicates the state of the $\emph{i}$th sample. In this paper, six affine parameters compose the state $x_t=[d_1,d_2,\theta_t,s_t,\alpha_t,\phi_t]$, where $d_1,d_2,\theta_t,s_t,\alpha_t,\phi_t$ represent the translation along horizontal and vertical direction, rotation angle, scale, aspect ratio and skew respectively. The random walk is employed as the transition model for $p(x_t|x_{t-1})$ i.e. $p(x_t|x_{t-1})=N(x_t;x_{t-1},\Psi)$, $N$ denotes a Gaussian distribution. $\Psi$ stands for the diagonal covariance matrix containing standard deviation of the six affine parameters in $x_t$.
\section{Saliency Guided Global Search}
\begin{figure*}[thpb]
  % Requires \usepackage{graphicx}
  \centering
\subfigure {\includegraphics[width=0.9\hsize]{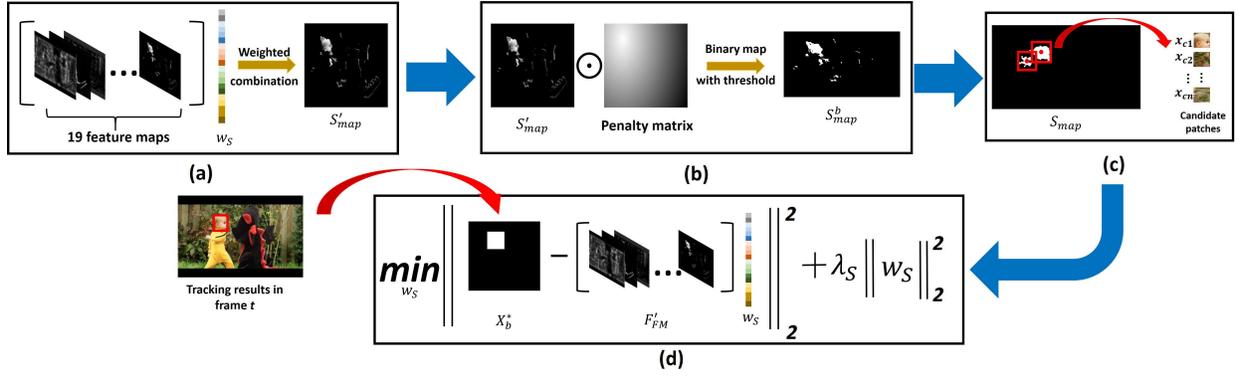}}
  \caption{Pipeline of saliency guided global search. (a) Linearly combine 19 feature maps with pre-determined weights $w_S$. (b) Penalize the combined map with estimated target position in last frame and generate binary saliency map. (c) Find the maximum connected area and determine the geometric center for candidate patch generation. (d) Update the weights $w_S$ with the tracking result in current frame.} \label{pipsal}
\end{figure*}

Traditional bottom-up attention model is not suitable for visual tracking since it fails to provide salient maps on the objects that we are interested in. To handle this, a novel top-down visual attention model is proposed to learn the saliency map that can emphasize the tracking target. The final saliency map is constructed through the combination of nineteen low-level features with different weights. The weights are updated based on the tracking result of the current frame. The outputs of this global search (candidate particles) are generated by calculating the connected area on the saliency map. They act as the samples of potential target position and passed into the integration mechanism. The overall pipeline of the global search can be described in Figure \ref{pipsal}. The saliency guided search will efficiently cope with challenging scenarios such as abrupt motion, motion blur as well as the out-of-plane rotation.\\
\subsection{Attention Map Building and Candidate Particle Generation}
In saliency detection, different levels of feature are usually utilized to construct a saliency map. 33 features distributed from low-, mid- and high-level are computed as the input for a SVM classifier in \cite{judd2009learning}. But the high-level features which detect a person or a face is not applicable since they need off-line training and tracking targets are not restricted to humans. Similar situation holds for the mid-level feature which assumes salient objects mostly lie on a horizon place. Thus, in this paper, we take advantage of the low-level features that can depict the fundamental and general characteristics of the object.\\
\indent By considering the tradeoff between validity and speed, nineteen low-level features are exacted firstly on a newly arrived frame. Similar to \cite{judd2009learning}, steerable pyramid subbands \cite{simoncelli1995steerable} in four orientation and three scales form the first 13 feature maps denoted as $F_{SPi}$, $i=1...13$. We also incorporate four broadly tuned color channels ($F_R, F_G, F_B, F_Y$) as well as the intensity channel ($F_I$) \cite{itti1998model} which are created by
\begin{eqnarray}
\notag F_R=r-(g+b)/2, F_G=g-(r+b)/2, F_B=b-(r+g)/2,
\end{eqnarray}
\begin{eqnarray}
F_I=(r+g+b)/3, \quad F_Y=(r+g)/2-|r-g|/2-b
\end{eqnarray}
where $r, g, b$ are red, green and blue channel of the image. Considering that the tracker is very likely to track humans, a skin color channel is involved as $F_{SK}$. The concatenation of the 19 feature maps construct the candidate feature map set $F_{FM}=[F_{SP1},...,F_{SP13},F_R, F_G, F_B, F_Y, F_I, F_{SK}]$. To release computational burden, similar to \cite{judd2009learning}, the original image is warped into $200\times200$ for feature map computation. Define a weight vector $w_{S}=[w_{S1},w_{S2}...w_{S19}]$ indicating the correlation degree between individual feature map and the tracking target. The method to determine and update $w_{S}$ will be introduced in next subsection. A top-down saliency map is created through the weighted sum of the low-level feature maps as
\begin{eqnarray}
\label{salp}
S^{'}_{map}=\sum_{i=1}^{19} w_{Si}F_{FMi}
\end{eqnarray}
where $F_{FMi}$ is the \emph{i}th element of $F_{FM}$. To incorporate the position information in the last frame into the saliency map, a revised center prior penalization is introduced in this work. Different from the traditional center prior \cite{judd2009learning} which believes human naturally tend to frame the object of interest near the center of the image, we penalize the saliency value in $S^{'}_{map}$ with the distance to the center of target in the last frame as $S^{c}_{map}=C(p_c)\bigodot S^{'}_{map}$, where $\bigodot$ is the Hadamard product (element-wise product). $C(p_c)$ represents the penalty matrix defined as
\begin{eqnarray}
\label{tarpen}
C_{ij}(p_c)=\delta_s\displaystyle\frac{Dis(p_{ij},p_c)}{\max{Dis(p_{ij},p_c)}}
\end{eqnarray}
where $p_c$ and $p_{ij}$ denote the center point of the target in the last frame and a certain point on $S^{'}_{map}$. $Dis(p_{ij},p_c)$ returns the Euclidean distance between the two points. $\delta_s$ is a tunable scaler and set as 2 in this paper.
\begin{remark}
Proper selection of $\delta_s$ is of importance in eliminating the false salient regions as well as reserving the true target for abrupt motion handling. Too small value is not adequate for the suppression of disturbance. On the contrary, too large value will lead to the points far from $p_c$ vanish severely which may also block the true target when abrupt motion occurs.
\end{remark}

\indent For the purpose of noise attenuation and easy operation, the distance penalized saliency map $S^{c}_{map}$ is subject to binarization with a threshold $\delta_b$ to produce the final saliency map $S_{map}$. The procedure introduced by now is described in Figure \ref{pipsal} (a) (b).\\
\indent In this paper, we assume the tracking target or part of it can map to a connected area on the saliency map $S_{map}$. The intuition behind this assumption is that the parts on an object which possess features with high distinction degree are usually coherent, for example a face with shin color against the entire head with black hair and red mouth. With this assumption, an ``run-relabel" algorithm \cite{haralock1991computer} is employed to derive connected area set $\Omega_c=[\omega_{c1}, \omega_{c2},...,\omega_{cn}]$ with a predefined threshold $\sigma_s$. $\sigma_s$ controls the minimum area that can be selected into set $\Omega_c$. Subsequently, we return the center location of each area to form $C_c$. We apply the same size and orientation of the bounding box in the last frame to every center element in $C_c$ to crop out the candidate particles $X_{c}=[x_{c1}, x_{c2},...,x_{cn}]$ as shown in Figure \ref{pipsal} (c). The particles are then passed into a judgement system for further evaluation which will be elaborated in ``Integration Mechanism" section. \\
\subsection{Online Weight Updating}
\begin{figure}[thpb]
  % Requires \usepackage{graphicx}
  \centering
\subfigure {\includegraphics[width=1\hsize]{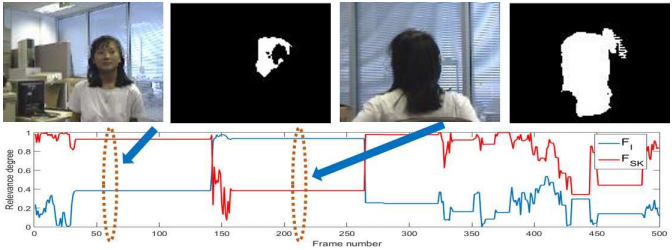}}
  \caption{Upper row: Images and generated saliency maps on \emph{Girl} of frame 65 and 210. Lower row: Relevance degree variation of feature $F_I$ and $F_{SK}$.} \label{girlupdate}
\end{figure}

In order to adapt to the appearance changes of the object and enhance the discriminative power, the weight $w_S$ needs to be updated dynamically. The update utilizes the prior information of the current frame after the target is estimated. Firstly, a binary groundtruth map $X_{b}^{*}(x,y)$ is built as follow
\begin{eqnarray}
X_{b}^{*}(x,y)=\begin{cases}1, \quad (x,y) \in \phi_b\\
0, \quad otherwise \end{cases}
\end{eqnarray}
where $\phi_b$ is a pixel set indicating the estimated bounding box in current frame. Together with $X_{b}^{*}$, the weights will be updated through an $L_2$ optimization problem.
\begin{eqnarray}
\label{wS}
\min_{w_{S}}\|X_{b}^{*}-F_{FM}^{'}w_{S}\|_{2}^2+\lambda_{S}\|w_{S}\|^2_{2}
\end{eqnarray}
where $F_{FM}^{'}$ is the vectorized version of candidate feature set $F_{FM}$.  $\lambda_S$ is a penalty parameter. The optimal solution to (\ref{wS}) is computed as
\begin{eqnarray}
\label{wsupdate}
w_{S}^{*}=(F^{'T}_{FM}F^{'}_{FM}+\lambda_{S}I)^{-1}F^{'T}_{FM}X_{b}^{*}
\end{eqnarray}
In order to avoid the error accumulation, the weight update is only conducted when a certain evaluation criterion is met instead of updating for every frame. The details can be found in the following section. For further clarification, a case study on dataset of \emph{Girl} is carried out as Figure \ref{girlupdate} shows. The original images of frame 65, 210 are extracted and the corresponding binary saliency maps are presented in the upper row. The relevance of two representative feature maps (intensity, skin color) to the final saliency map is investigated and plotted as the lower row shows. Considering the fact that the relevance increases if the corresponding weight approaches to 1, the relevance degree $R(w_{Si})$ of individual feature can be quantitatively computed as
\begin{eqnarray}
R(w_{Si})=\begin{cases}-w_{Si}(w_{Si}-2), & w_{Si}\leq1\\ \displaystyle\frac{1}{e^{w_{Si}-1}} &w_{Si}>1\end{cases}
\end{eqnarray}
From the study, it can be observed that the binary saliency map can efficiently predict the rough location of the target even if severe appearance changes occur. Combining the relevance degree curves, we can find that the skin color feature dominates over other ones before the girl turns back. When she shows her hair towards the camera, the intensity feature becomes the main descriptor for the saliency map construction. This switching can be efficiently handled by the online weight update. It should be noted that although the saliency map cannot provide accurate silhouette of the object, it is adequate to act as a coarse search result that indicates the rough location. A fine search procedure is required to achieve a robust and accurate tracking as introduced in local search section.
\section{Integration Mechanism}
The integration mechanism introduced in this paper performs as an evaluation process for the generated candidate particles $X_c$ from global search. Due to the possible clutter background or identical fake targets in video sequence, the outcome particles from global search may also contain false target patches. Thus, an algorithm with strong local discriminative power is required to rank the particles. By considering the balance between accuracy and speed, the L2-RLS \cite{xiao2014l2} tracker with PCA and square template $D=[B,E]$ is employed with the following likelihood function
\begin{eqnarray}
\label{confi}
p(y_i|x_{ci}) \propto \exp(-\|y_i-D\beta\|^2_2-\delta_c\|e_e\|_1)
\end{eqnarray}
where $\beta=[\beta_b,\beta_e]$ denotes the sparse coefficient and $e_e=E\beta_e$. $\delta_c$ is a tunable scalar. The confidence value of each candidate particle denoted as $\Theta_c=[\theta_{c1}, \theta_{c2},...,\theta_{cn}]$ is calculated with (\ref{confi}). Then, the maximum confidence is derived as $\theta_c^{*}=\max \{\theta_{ci}\}$ and it is regarded as the most promising candidate patch that enclose the target.\\
\indent To achieve the evaluation, we set two thresholds $\tau_c<\bar{\tau}_c$ for the maximum confidence $\theta_c^{*}$ and three cases are considered. (i) If $\theta_c^{*}>\bar{\tau}_c$, it indicates that the global search has already provided a sufficiently accurate location prediction, hence, the local search will be ignored directly and the corresponding patch $x^{*}_c=\mathop{\argmax}_{x_{ci}}{\{\theta_{ci}\}}$ is regarded as the target. (ii) If $\tau_c<\theta_c^{*}<\bar{\tau}_c$, this means the output is acceptable but not accurate. Thus, a further local search is required centered at the geometric center of patch $x_c^{*}$. This case may occur because the saliency map $S_{map}$ after binarilzation and connected area thresholding would only present a certain salient part of the target. In this way, there may exist displacement between the geometric center of the extracted connected area and the actual target center. (iii) If we have $\theta_c^{*}<\tau_c$, the output of global search is very likely to have drifted away due to the possible cluster background or illumination variation. In this case, local search is performed normally centered at the target location in last frame.\\
\indent Additionally, an extra threshold $\tau_c<\tau_{cw}<\bar{\tau}_c$ is needed as the weight update judgement criterion for $w_S$ updating. If $\theta_c^{*}$ is larger than $\tau_{cw}$, we can tell that the current image condition is suitable for weight updating. Otherwise, the image is undergoing severe external interference which may lead to the drift of the entire tracker. To equip the proposed tracker with re-initialization power after drifting, the weight update suspends in this case. When the external interference factor vanishes, the global search will be able to guide the tracker to capture the target again.
\section{Local Search with Hierarchical Structure }
\subsection{Superpixel Matching via HSV Histogram}
\begin{figure}[thpb]
  % Requires \usepackage{graphicx}
  \centering
\subfigure {\includegraphics[width=1\hsize]{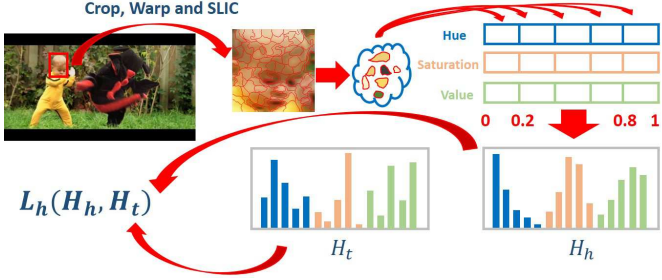}}
  \caption{Flowchart of superpixel HSV matching} \label{supflow}
\end{figure}

Most of the current appearance template matching algorithm depends only on the target intensity for template building. In this way, the color information is neglected and the accuracy degrades since the color characteristics provide extra discriminative power to distinguish the target and the background. To efficiently describe the color, compared with low-level pixel-wise cue, the superpixels as a mid-level cue presents color structural information about the details of an object. Thus, in this paper, a novel superpixel matching is proposed as an enhancement to sole intensity template matching by investigating the color distribution of the target object. This method can be summarized as Figure \ref{supflow}.\\
\indent The superpixel matching is still embeded into the L2 tracker mentioned in the last section to achieve a collaborative tracking. $N_s$ patches with the highest confidence are passed to the superpixel matching process. There are two main reasons for employing this structure: (i) the generation of superpixel segments is time-consuming, we can accelerate the algorithm by reducing the search set and (ii) it is challenging for the proposed superpixel matching method to handle severe occlusion and illumination changes. On the contrary, together with square template, L2 tracker has excellent capability in coping with these problems. Thus, incorporating with L2 tracker can complement the drawbacks of superpixel matching approach. \\
\indent Each image patch from the $N_s$ selected particles is firstly subject to segmentation to generate $N_h$ superpixels $S=[s_1, s_2,...,s_{N_h}]$ using SLIC \cite{achanta2012slic}. The superpixel segments are then converted to a normalized HSV color space before a five-bin histogram is created for each channel. The length of each bin is evenly allocated on the full scale of 0-1. The three histograms are denoted as $H_o$ with $(o=H,S,V)$ and computed as
\begin{eqnarray}
\label{hisbin}
H_{oj}=\sum_{i=1}^{N_h}\exp{[-k_o(s_{oi}-c_{oj})^2]}
\end{eqnarray}
where $H_{oj}$ is the \emph{j}th bin of histogram $H_o$ and $c_{oj}$ denotes the center of individual bin. $k_o$ is a scalar chosen as 10. The expression in (\ref{hisbin}) measures the distance between each superpixel patch and the bin centers under different channels which reflects the color distribution. Subsequently $H_{o}$ is concatenated to yield $H_{h}$ as the figure shows. The generated HSV histogram $H_{h}$ will be compared with a template $H_t$. This template is initialized with the target patch in the first frame and updated with the following update law.
\begin{eqnarray}
\label{htupdate}
H_t=\gamma H_t+(1-\gamma)H^{*}_{h}
\end{eqnarray}
where $\gamma$ is a learning rate which is tuned as 0.95 in this paper. $H^{*}_{h}$ denotes the optimum histogram that possesses the maximum joint observation likelihood which will be introduced in the following context. We then introduce a cosine similarity to measure the similarity between $H_h$ and the template histogram $H_t$.
\begin{eqnarray}
L_h(H_{h},H_t)=\displaystyle\frac{\langle H_{h},H_t\rangle}{\|H_{h}\|\|H_{t}\|}
\end{eqnarray}
The similarity $L_h(H_{h},H_t)$ is then transformed into an error $e_h=\frac{1}{k_h+L_h}$ pattern where $k_h$ is a positive parameter. Following the idea of multi-objective optimization \cite{miettinen2012nonlinear}, the reconstruction error $e_b=y-B\beta_b$ from the L2 tracker and HSV histogram matching error $e_h$ are fused to produce the confidence for each particles. Both of the errors are concatenated into a new vector $e_v=[e_{v1},e_{v2}]^T$ where $e_{v1}=e_a$ and $e_{v2}=e_h$. The ideal point $e_v^*=[\min\{e_{v1}^i\},\min\{e_{v2}^i\}]^T$ is derived where $i$ denotes the index of candidate particles. Then the joint observation likelihood of state $x_i$ is calculated as follow.
\begin{align}
\label{condce}
\notag p(y_i|x_i)\propto &\exp[-\mu_1(e_{v1}^i-e^*_{v1})/(\max\{e_{v1}^i\}-e^*_{v1})\\
&-\mu_2(e_{v2}^i-e^*_{v2})/(\max\{e_{v2}^i\}-e^*_{v2})]
\end{align}
where $\mu_1+\mu_2=1$ are used to adjust the ratio between these two cues. Top $N_l$ particles with maximum observation likelihood are reserved and proceed to linear refinement search.
\subsection{Linear Refinement Search}
Due to the inevitable existence of inaccurate appearance template $B$, imperfect dictionary update scheme and drawbacks in superpixel histogram matching, the estimated result computed by traditional MAP may not be so reliable since only the one that has the maximum confidence value is selected as the result. But, it is very likely that the most accurate candidate has lower confidence, thus, cannot contribute to the final result. To ameliorate this, we propose a novel linear refinement approach to balance the result by also incorporating the candidates that possess high confidence into the final coding. \\
\indent In refinement search phase, $N_l$ particles with the highest confidence are extracted to achieve a linear combination for the final estimation. The particles construct a new candidate $M=[x^*_1,x^*_2,...,x^*_{N_l}]$. Then we solve the following optimization problem.
\begin{eqnarray}
\label{refin}
\{\alpha,\beta\}=\arg \min_{\alpha,\beta}\|M\alpha-D\beta\|_2^2+\kappa\|\beta\|_2^2, s.t. \sum_{i=1}^{N_l}\alpha=1
\end{eqnarray}
where $\alpha$ and $\beta$ are the coefficients for candidate set $M$ and the PCA dictionary $D$. The intuition behind this approach is that we hope to use the linear combination of the candidate patches to reconstruct the target instead of a single one to upgrade the accuracy. This idea is similar to \cite{wang2015visual}. The main differences can be highlighted as (i) our method only consider the $N_l$ particles with high confidence value instead of the entire ones in order to reduce the possibility that the final result is interfered by ``bad" particles; (ii) since in practical implementation, $N_l$ is usually not larger than 10, the sparse constraint for $\alpha$ can be eliminated in (\ref{refin}). In this way, by still adopting the $L_2$ norm constraint for $\beta$, an iterative analytical solution to the optimization problem can be found. The computational burden can be hugely released.\\
\indent The coefficients $\alpha$ and $\beta$ can be solved iteratively by fixing one and updating the other one. Given fixed $\alpha^{(i)}$ where $i$ represents the \emph{i}th iteration, and consequently $\hat{y}=M\alpha^{(i)}$, solving $\beta^{(i+1)}$ becomes a $L_2$ optimiztion problem which gives
\begin{eqnarray}
\label{betasol}
\beta^{(i+1)}=(D^TD+\kappa I)^{-1}D^T\hat{y}
\end{eqnarray}
Then we fix $\beta^{(i)}$ which yields $\hat{y}=D\beta^{(i)}$, the original problem becomes a linearly-constrained minimum Euclidean norm problem which has an analytical solution as \cite{liu2014spatial}
\begin{align}
\label{alphsol}
\notag \alpha^{(i+1)}=&(M^TM)^{-1}\{[I-\frac{1}{l^T(M^TM)^{-1}l}ll^T(M^TM)^{-1}]\\
&M^T\hat{y}+\frac{1}{l^T(M^TM)^{-1}l}l\}
\end{align}
where $l$ denotes a vector with all ones. The iteratively solving program terminates when a stopping criterion is satisfied and output a $\alpha^{(n)}$. The optimum coefficient $\alpha^*$ is derived by revising the negative entries in $\alpha^{(n)}$ to zero followed by a normalization operation. Finally, the estimated target $\hat{x}$ is computed by linearly combining the $N_l$ particles based on $\alpha^*$, i.e.
\begin{eqnarray}
\hat{x}=\sum_{i=1}^{N_l} \alpha^*_i x_i^*
\end{eqnarray}
The summary of the solver can be found in Algorithm \ref{iteslv}.
\begin{remark}
\label{remark1}
The iterative operation for solving (\ref{refin}) is efficient since both phases (\ref{betasol}) (\ref{alphsol}) have analytical solution. Moreover, (\ref{betasol}) and (\ref{alphsol}) can be rewritten as $\beta^{(i+1)}=F_{\beta}\hat{y}$ and $\alpha^{(i+1)}=F_{\alpha}\hat{y}+g_{\alpha}$ where $F_{\beta}=(D^TD+\kappa I)^{-1}D^T$, $F_{\alpha}=(M^TM)^{-1}\{[I-\frac{1}{l^T(M^TM)^{-1}l}ll^T(M^TM)^{-1}]M^T\}$ and $g_{\alpha}=(M^TM)^{-1}\frac{1}{l^T(M^TM)^{-1}l}l$. Throughout the solving process, the new candidate set $M$ and the dictionary $D$ are fixed which means $F_{\alpha}$, $F_{\beta}$ and $g_{\alpha}$ remain unchanged and can be computed prior to the calculation loop. Therefore, the computational complexity can be dramatically reduced.
\end{remark}
\begin{algorithm}
  \caption{Iterative solver for (\ref{refin})}
  \label{iteslv}
  \begin{algorithmic}[1]
  \STATE  \textbf{Initialization:} Initialize $\alpha^{(0)}$ with corresponding confidence derived from (\ref{condce}) and normalize it to guarantee $\sum_{k=1}^{N_l}\alpha^{(0)}_k=1$. Calculate $F_{\alpha}$, $F_{\beta}$ and $g_{\alpha}$ according to the description in Remark \ref{remark1} using $M$ which contains $N_l$ particles with highest confidence and the appearance template $D$.
  \WHILE{Neither the solution is convergent nor the maximum iteration number is met}
 \STATE   $\hat{y}=M\alpha^{(i)}$\\
   \STATE $\beta^{(i+1)}=F_{\beta}\hat{y}$\\
  \STATE $\hat{y}=D\beta^{(i+1)}$\\
  \STATE $\alpha^{(i+1)}=F_{\alpha}\hat{y}+g_{\alpha}$\\
  \STATE i=i+1
  \ENDWHILE
  \STATE Vanish the negative entries and normalize $\alpha^{(n)}$ to produce $\alpha^{*}$
  \end{algorithmic}
\end{algorithm}

\begin{figure}[thpb]
  % Requires \usepackage{graphicx}
  \centering
\subfigure {\includegraphics[width=0.8\hsize]{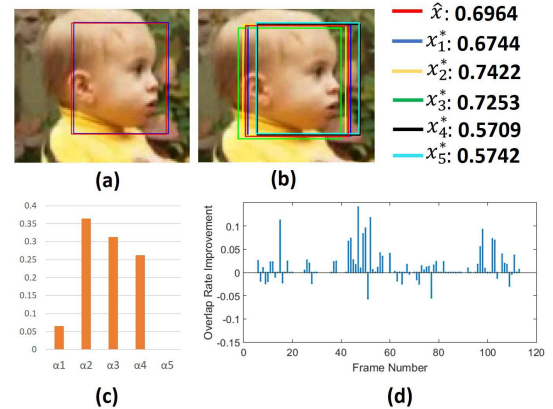}}
  \caption{Case study on frame 11 on \emph{DragonBaby}. $N_l=5$ (a) The bounding box of refined result $\hat{x}$ and candidate $x_1^{*}$ with highest confidence. (b) Bounding boxes of $\hat{x}$ and all the candidates $x^{*}_1$-$x^{*}_5$. (c) Coefficients $\alpha$ of five candidates. (d) Overlap rate improvement between the refined results $\hat{x}$ and $x_{1}^{*}$ throughout the sequence.} \label{refevl}
\end{figure}

\indent A case study on the video sequence of \emph{DragonBaby} is conduced to evaluate the performance of refinement approach. The result is shown as Figure \ref{refevl}. From (a), we can see that the bounding box of the refined result $\hat{x}$ (red box) shows higher overlap rate to the real target than the one with highest confidence $x^{*}_1$ (blue box). The quantitative study can also support this observation (0.6964 vs 0.6744). The result of all the five candidates and the corresponding overlap rate is presented as (b). It should be noted that the candidate $x^{*}_2$ with the second highest confidence value possesses the largest overlap rate followed by $x^{*}_3$. Although the overlap rate of refined result is not competitive to $x^{*}_2$ and $x^{*}_3$, it improves compared to $x^{*}_1$ which would be regarded as the outcome in MAP. Besides, the weight distribution of the five candidates is shown in (c), $\alpha_2$ and $\alpha_3$ have high weights corresponding to $x^{*}_2$ and $x^{*}_3$ which indicate that they contribute much to the final result and lead to an improved accuracy. However, due to the imperfect template $D$ and other disturbance, $x^{*}_4$ with lowest overlap rate is also assigned with a relatively high weight. This phenomenon is difficult to avoid since we are lack of sufficiently accurate priori knowledge to judge a candidate before the refinement and get rid of these ``bad" candidates. But by involving the refinement approach, the output is achieved in a balanced way and the effect of ``bad" candidates can be attenuated. A study to measure the improvement from the proposed refinement approach to the traditional MAP is carried out on the entire sequence and the result is plotted as (d). We ignore the frames that skip local search according to the integration mechanism. The figure is plotted by subtracting the overlap rate of $x^{*}_1$ from the refined result $\hat{x}$. It can be observed that although, for the minority of frames, the refinement approach degrades the accuracy, for most frames, it can level-up the performance. And the maximum improvement scale can be as large as 0.15.\\
\indent The proposed algorithm in this paper can be summarized by Algorithm \ref{overallsum}.
\begin{algorithm}
  \caption{The summary of proposed SHT tracker}
  \label{overallsum}
  \begin{algorithmic}[1]
  \STATE  \textbf{Initialization:} Initialize weight vector $w_S$, PCA and square template $D$, HSV histogram template $H_t$ and other coefficients.
  \STATE \textbf{Input:} Current frame $f_t$
  \STATE \textbf{Output:} Estimated target location $\hat{x}$
  \STATE \textbf{Start:} Resize the image to $200\times200$ and generate 19 low-level feature maps $F_{FM}$.
  \STATE Produce the saliency map $S_{map}$ through (\ref{salp}), target location penalization (\ref{tarpen})and binarization.
  \STATE Find the centers $C_c$ of connected regions that have larger areas than threshold $\sigma_s$.
  \STATE Crop the candidate particles $X_c$ with the same size and orientation of bounding box in $f_{t-1}$.
  \STATE Calculate corresponding confidence for $X_c$ using (\ref{confi}) and pick the maximum one $\theta_c^*$.
  \IF {$\theta_c^{*} > \bar{\tau}_c$}
  \STATE Skip local search and set $\hat{x}=x_c^{*}$
  \ELSIF {$\tau_c<\theta_c^{*}<\bar{\tau}_c$}
  \STATE Perform local search by firstly sampling at the geometric center of patch $\theta_c^{*}$.
  \STATE Apply $L2$ tracker as described in \cite{xiao2014l2} and extract $N_s$ particles with highest confidence values.
  \STATE Generate HSV histogram $H_{oj}$ for each patch using (\ref{hisbin}) and concatenate them into $H_h$.
  \STATE Compute the similarity $L_h(H_h,H_t)$ and the corresponding error $e_h$.
  \STATE Calculate the joint observation likelihood with (\ref{condce}) and proceed to linear refinement search with $N_l$ patches that have highest confidence values to form $M$.
  \STATE Solve the optimization problem (\ref{refin}) with Algorithm \ref{iteslv} to derive $\alpha^{*}$.
  \STATE Obtain refined result as $\hat{x}=\sum_{i=1}^{N_l} \alpha^*_i x_i^*$
  \ELSE
  \STATE Perform the local search using the particles sampled at the center of estimated target location in $f_{t-1}$.
  \ENDIF
  \STATE Update dictionary $D$ as depicted in \cite{xiao2014l2} and update $H_t$ using (\ref{htupdate}).
  \IF {$\theta_c^{*} > \tau_{cw}$}
  \STATE Update $w_S$ with the groundtruth map $X_b^*$ using (\ref{wsupdate}).
  \ENDIF
  \end{algorithmic}
\end{algorithm}
\section{Experiments}
The proposed SHT algorithm is implemented in MATLAB and run on an Intel Core i7-4710HQ 2.5GHz PC with 16GB memory. The running speed is at around 1.5 FPS without any code optimization. The number of particles $N_s$ and $N_{l}$ for superpixel matching and refinement search is 70 and $5\sim10$. The regularization parameters $\lambda_S$, $\kappa$ in update law (\ref{wS}) and refinement search (\ref{refin}) are selected as 0.05 and 0.005 respectively. The thresholds for the availability of saliency-guided candidate patches $\tau_c$ and $\bar{\tau}_c$ need to be tuned for each dataset. Experientially, they belong to the range of [0.2, 0.45] and [0.4, 0.8]. The template size in $L2$ tracker remains to be $32\times32$ and the number of PCA templates in $B$ is 16. 600 particles are sampled for each frame. The test is conducted both qualitatively and quantitatively on ten challenging video sequences. And the results are compared with eleven state-of-the-art trackers including L1APG \cite{bao2012real}, IVT \cite{ross2008incremental}, Frag \cite{adam2006robust}, TLD \cite{kalal2010pn}, CSK \cite{henriques2012exploiting}, L2-RLS \cite{xiao2014l2}, ASLA \cite{jia2012visual}, SCM \cite{zhong2014robust}, CXT \cite{dinh2011context}, LOT \cite{oron2015locally}, DFT \cite{sevilla2012distribution}.
\subsection{Component Validation}
\begin{figure}[thpb]
  % Requires \usepackage{graphicx}
  \centering
\subfigure {\includegraphics[width=1\hsize]{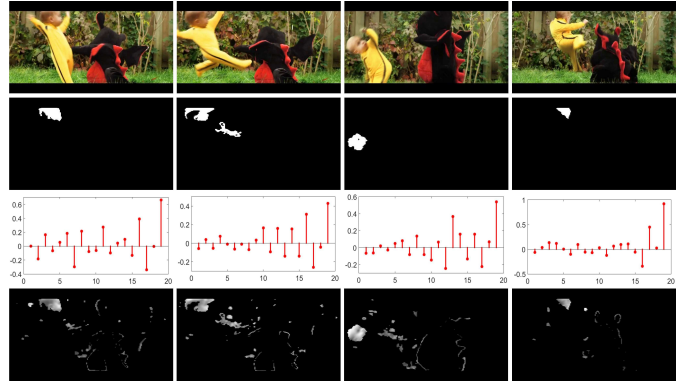}}
  \caption{Case study on \emph{Dragonbabdy} for saliency guided search. First row shows the original images of frame \emph{42}, \emph{44}, \emph{46} and \emph{51}. Second row presents the binary saliency maps. Third row is the corresponding weights to the 19 feature maps. Last row is the feature map of skin-color channel.} \label{DBSal}
\end{figure}

\begin{figure}[thpb]
  % Requires \usepackage{graphicx}
  \centering
\subfigure {\includegraphics[width=1\hsize]{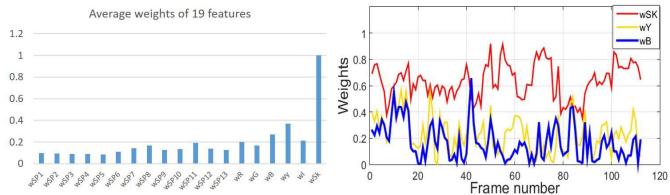}}
  \caption{Histogram of accumulated feature weights and the weight updating of three most salient ones.} \label{salweivaria}
\end{figure}

This section demonstrates the effectiveness and efficacy of the three main components of SHT: saliency guided global search, superpixel matching and linear refinement search. Figure \ref{DBSal} shows a case study on \emph{DragonBaby} dataset for the global search. Four frames with severely abrupt motion and incomplete target are extracted on the first row. The second row presents the final binary saliency map $S_{map}$ which can roughly capture the location of the target. Although, there is still a false saliency area after the binarization and connected area thresholding on the second map, the integration mechanism can facilitate to abandon it and guide the tracker to capture the correct target as shown in Figure \ref{DeeDuk1}. The third row demonstrates the weight allocation of the 19 feature maps $F_{FM}$ in the four frames. It can be noted that the last feature $F_{SK}$ which describes the skin color is always assigned with highest weight. This observation is consistent with the fact that we are tracking a human head. The dominant role of this feature can be further proven by the histogram of accumulated feature weight in Figure \ref{salweivaria}. This histogram represents the normalized summation of the absolute value of each weight. The yellow color feature map $F_Y$ occupies the second high accumulated weight. The first reason is straightforward that yellow is the most identical color to human skin. However, if we look further to the weight allocation plot in Figure \ref{DBSal}, it can be found that the sign of the $w_y$ is mostly negative (first three frames) which can be utilized to counteract the false saliency regions in $F_{SK} $which have similar yellow color as shown in the last row of the figure. These false regions are caused by the background objects with similar color channel such as the leaves. In this way, together with the binarization and connected area thresholding, a pure saliency map can be achieved. Finally, the weight updating curves as shown in Figure \ref{salweivaria} show that the weights are updated per frame in this case because there is no severe occlusion occurs and every frame is suitable for updating.\\

\begin{figure*}[thpb]
  % Requires \usepackage{graphicx}
  \centering
\subfigure {\includegraphics[width=1\hsize]{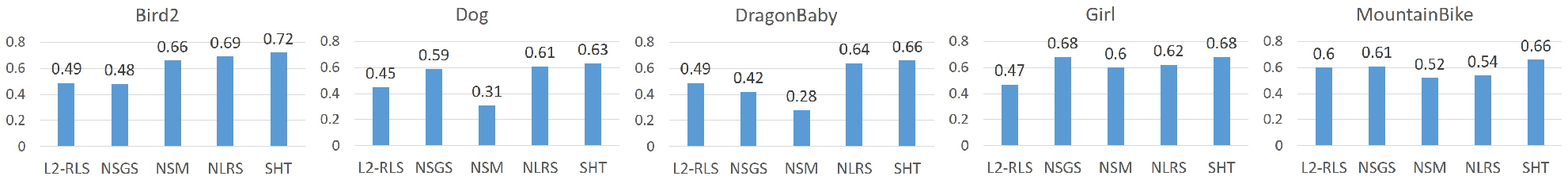} \label{com1} }
\subfigure {\includegraphics[width=1\hsize]{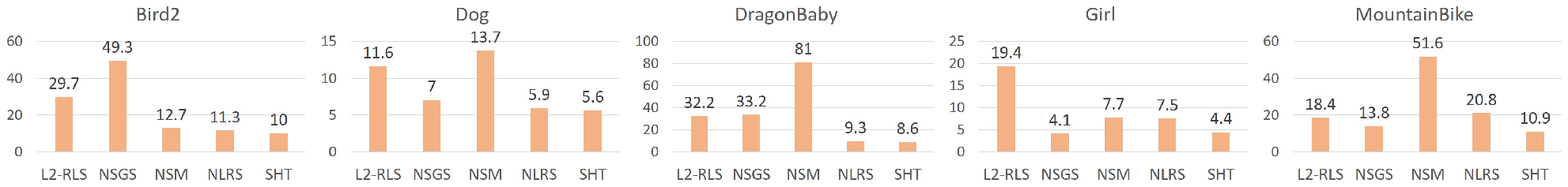} \label{com2} }
  \caption{Tracking results of individual component. Blue and yellow bins denote the overlap rate and center error for each data set. \emph{L2-RLS}, \emph{NSGS}, \emph{NSM}, \emph{NLRS} and \emph{SHT} mean the sole appearance matching in \cite{xiao2014l2},  No Saliency Guided Search, No Superpixel Matching, No Linear Refinement Search and the overall proposed tracker.} \label{comval}
\end{figure*}
\indent For the purpose of thorough study, additional investigation is performed on five datasets in the absence of a certain component. The result is shown as Figure \ref{comval}. In \emph{NSGS} study, we block the saliency module as well as the integration mechanism and perform the particle sampling centered at the location of target from the last frame. For \emph{NLRS}, the target is predicted with traditional MAP that maximize the observation likelihood (\ref{condce}). Generally, the proposed SHT algorithm can outperform sole L2 tracker in some challenging scenarios such as out-of-plane rotation (\emph{Bird2}, \emph{Girl}), deformation (\emph{Dog}, \emph{MountainBike}), abrupt motion (\emph{DragonBaby}). Thanks to the global search ability of the saliency guided approach, an obvious performance level-up can be achieved between \emph{NSGS} and \emph{SHT} on the datasets with discriminative target appearance and the background (\emph{Bird2}, \emph{DragonBaby}). Although, due to the weight updating design, a pure and available saliency map can be obtained for \emph{Girl}, the accuracy is not improved much. This is because the face motion of the girl is very slow, hence, a proper standard deviation in particle sampling can sufficiently cover the range of the motion. As for the superpixel matching, it is especially efficient for the cases that have consistent color distribution. For example, the target in \emph{Dog} undergoes severe deformation when the dog running towards the woman. But the color composition (black and white) remains unchanged which is very suitable for the superpixel matching. The same situation can be observed for the case of \emph{Bird2} and \emph{MountainBike}. Besides, even if the color composition of the target changes, the template updating mechanism can still guarantee the robustness of the tracker as demonstrated in \emph{Girl}. From \emph{NLRS}, we can judge that the refinement approach will slightly improve the accuracy since essentially, it does not bring in any new cue for tracking. But due to its capability in attenuating the influence of ``bad" candidates and efficient computational time, the involvement of refinement an auxiliary procedure is beneficial for tracking .\\
\begin{table}[thpb]
\centering
\small
\caption{Running time of individual component}\label{comtime}
\begin{tabular}{c c c c c c c c c c c c c c c c c} % centered columns (4 columns)
    \hline\hline
       & SGS&SM&LRS&L2RLS&SHT \\ \hline\hline
       Cost time (s)&0.15&0.40&0.001&0.12&0.67\\ \hline
       \multicolumn{6}{l}{\tiny SGS: Saliency Guided Search. SM: Superpixel Matching. LRS: Linear Refinement Search.}
    \end{tabular}
\end{table}
\indent The time consumption of each component is listed in Table \ref{comtime}. As the table shows, superpixel matching is the most time-consuming part of SHT because of the SLIC process. The linear refinement almost doest not cost any time since analytical solution can be found for each iteration in the solver. Finally, the speed of feature extraction in SGS is not fast but acceptable.
\subsection{Qualitative Evaluation}
\begin{figure*}[thpb]
  % Requires \usepackage{graphicx}
  \centering
\subfigure [Selected frames of tracking results for Sequences \emph{Birds}, \emph{Boy}, \emph{Caviar} and \emph{Dog}.] {\includegraphics[width=1\hsize]{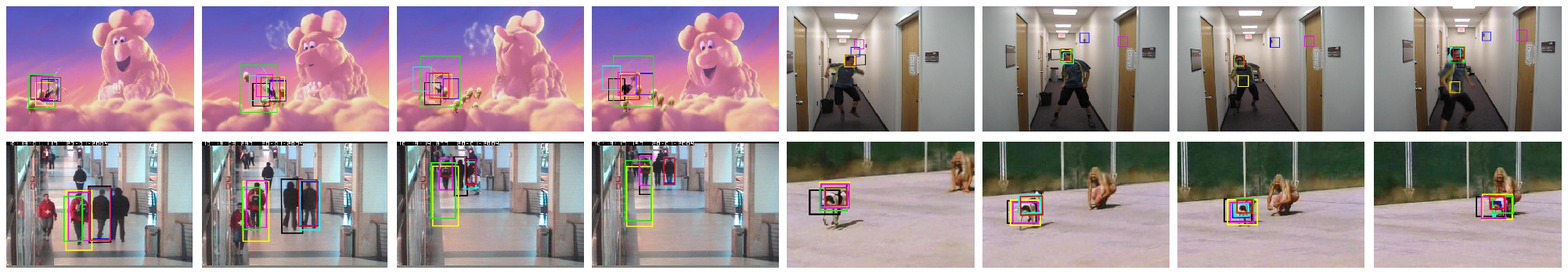} \label{DeeDuk} }
\subfigure [Selected frames of tracking results for Sequences \emph{DroganBaby}, \emph{Girl}, \emph{Jogging}, \emph{MountainBike}, \emph{Singer1} and \emph{Walking2}.] {\includegraphics[width=1\hsize]{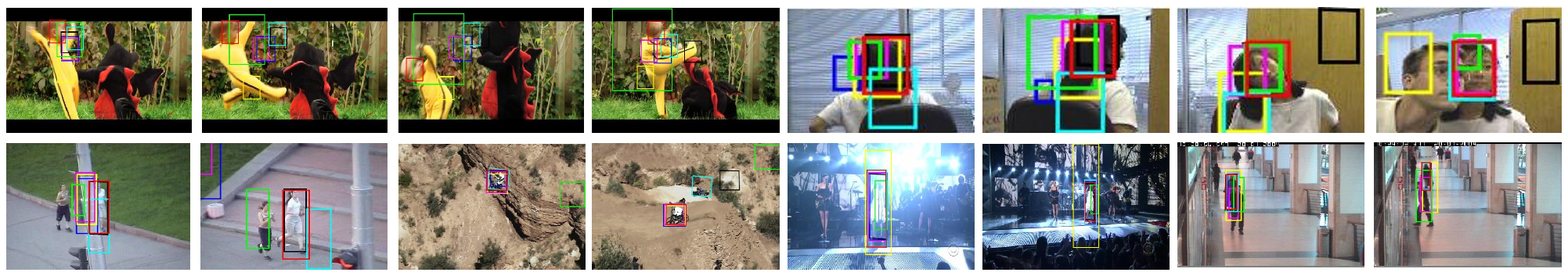}\label{DeeDuk1} }
\subfigure {\includegraphics[width=0.7\hsize]{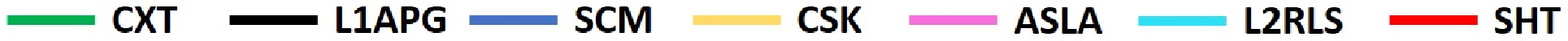}}
  \caption{ Tracking result screenshots of seven trackers. Video sequences of \emph{Caviar}, \emph{Girl}, \emph{Jogging}, \emph{Walking2} with heavy occlusions. Video sequences of \emph{Boy}, \emph{DragonBaby} with fast motion and motion blur. Video sequences of \emph{MountainBike}, \emph{Singer1} with background clutters and drastic illumination variation. Video sequences of \emph{bird2}, \emph{DragonBaby}, \emph{Girl}, \emph{MountainBike} with in-plane or out-of-plane rotation. Video sequences of \emph{Dog}, \emph{Jogging}, \emph{Bird2} with non-rigid object deformation.} \label{quastudy}
\end{figure*}
Qualitative investigation is conducted with ten challenging video sequences on six state-of-the-art trackers. The result is shown as Figure \ref{quastudy} and the analysis comes below.\\
\indent \textbf{Heavy occlusion}: Four representative datasets with partial or full occlusion are evaluated in this case. The capability for occlusion handling of SHT mainly inherits from the square template embedded in the L2 tracker. The ratio combination observation likelihood (\ref{condce}) ensures that the possible drift caused by the superpixel matching will not effect the performance when encountering occlusions. In \emph{Girl}, when the girl's face is partially blocked by the man in the last figure, only the proposed tracker and L2-RLS with square template and SCM, ASLA with part-based template can cope with this. Although, L1APG also employs trivial template and update scheme based on occlusion detection, the raw pixel dictionary makes the tracker susceptible to the cases when the target is blocked by similar objects as shown in the \emph{caviar} sequence.\\
\indent \textbf{Abrupt motion and motion blur}: It is usually very challenging for particle filter based tracker to deal with targets with fast motion due to the tradeoff between number of samples and sampling radius. Besides, the situation becomes more complicated since motion blur is always along with fast motion. The saliency guided global search can solve this problem favorably because it only focuses on salient regions in the whole image and provides a predicted location. Moreover, although motion blur brings challenges for template matching, its saliency feature remains the same, hence, saliency search is very robust to motion blur as well. The tracking result comparison of \emph{DragonBaby} is presented in Figure \ref{DeeDuk1}. The target undergoes fast motions in the first three figures which corresponds to frame 42,44,46. It can be observed that only SHT is able to complete the tracking. In the last figure, the baby turns back his head, but the color characteristics make it still discriminative to the background and the saliency search can successfully guide the tracking. In \emph{Boy}, together with the refinement procedure, the proposed method can provide a more accurate performance as shown in the second and fourth figure compared with sole L2 tracker. Apart from SHT, CXT also outperforms other trackers due to its essence of PN tracker and equipped with re-initialization mechanism. Moreover, the involvement of context information also contributes to its premium performance.\\
\indent \textbf{Background clutter and illumination variation}: these two situations are considered together because they both degrade the contrast between the foreground and background. In \emph{MountainBike}, the biker rides across the gap with varying postures. Also, the rocks and bushes inside the gap form clutter background that is very challenging for the saliency guided search since there may be many false saliency regions. Thanks to the integration mechanism, the false targets can always be filtered out and the tracker can capture the biker throughout the sequence. On the contrary, L2 tracker drifts to the rocks after the biker landing. This is because the greyscale holistic appearance template it uses is not sufficient to discriminate background and the target. The superpixel matching in SHT can handle this by investigating the color distribution of each particles. In \emph{Singer1}, the target undergoes drastic illumination changes under the state light as shown in Figure \ref{DeeDuk1}. All the sparse representation based trackers (L1APG, SCM, ASLA, L2, SHT) perform favorably due to the inherent robustness against illumination variation. CXT which depends on the context elements exploration is very sensitive to illumination variation and can not achieve the tracking when the light condition turns back to normal.\\
\indent \textbf{Rotation}: Four video sequences: \emph{Bird2}, \emph{DragonBaby},\emph{ Girl}, \emph{MountainBike} with in-plane and out-of-plane rotation are tested. In \emph{Bird2}, most of the trackers cannot handle the mirror rotation when the bird walks back, especially the deformation of the bird occurs at the same time. As analyzed in the last subsection, the color composition of the bird remains the same when the symmetric rotation happens. In this way, the superpixel matching can facilitate the task and rectify the incorrect output caused by the L2 tracker. In \emph{Girl}, the target is subject to complicated rotation, appearance changes and scale variation. Only SHT can adapt to the challenges at the same time.\\
\indent \textbf{Non-rigid object deformation}: Non-rigid deformation occurs when human or animal move and the appearance vary. In \emph{Dog} dataset, the superpixel matching can help to improve the performance as mentioned in component validation section. It should be noted that the shadow created by the woman is an interference factor especially when the dog shaking its tail under her and some trackers tend to enlarge the bounding box to enclose the shadow region due to the lack of robustness. In \emph{Jogging}, the deformation of the woman happens in a gentle way so that the proposed tracker have no problem in handling it. Most of the trackers are blocked by the pole at the beginning. Only L1APG and SHT complete the tracking.
\subsection{Quantitative Comparison}
\begin{table*}[thpb]
\centering
\small
\caption{Average overlap rate}\label{overlap}
\begin{tabular}{c c c c c c c c c c c c c c c c c} % centered columns (4 columns)
    \hline\hline
       & $L_1$APG & IVT & Frag & TLD & CSK & L2-RLS & ASLA& SCM& CXT&LOT &DFT&SHT \\ \hline\hline
       Bird2&0.20&0.49&0.43&0.27&\textcolor{green}{0.58}&0.49&0.50&0.46&0.24&0.09&\textcolor{blue}{0.61}&\textcolor{red}{0.72}\\ \hline
       Boy&0.51&0.25&0.45&0.62&\textcolor{green}{0.66}&\textcolor{blue}{0.78}&0.37&0.38&\textcolor{red}{0.80}&0.54&0.40&\textcolor{red}{0.80}\\
       \hline
       Caviar&\textcolor{green}{0.38}&0.15&0.26&0.20&0.14&\textcolor{blue}{0.86}&0.15&\textcolor{red}{0.87}&0.13&0.21&0.10&\textcolor{blue}{0.86}\\ \hline
        Dog&0.38&0.14&0.34&0.32&0.36&0.45&0.42&\textcolor{green}{0.56}&\textcolor{blue}{0.57}&0.47&0.32&\textcolor{red}{0.63}\\ \hline
        DragB&0.24&0.25&0.40&0.07&0.20&\textcolor{green}{0.49}&0.23&0.19&0.37&\textcolor{blue}{0.52}&0.14&\textcolor{red}{0.66}\\ \hline
        Girl& 0.42&0.17&0.47&0.49&0.38&0.47&\textcolor{blue}{0.65}&0.27&\textcolor{green}{0.61}&0.43&0.29&\textcolor{red}{0.68}\\ \hline
        Jogging&\textcolor{red}{0.83}&0.18&0.46&\textcolor{green}{0.72}&0.15&0.44&0.14&0.14&0.13&0.65&0.22&\textcolor{blue}{0.78}\\
        \hline
        MounB&0.62&\textcolor{red}{0.74}&0.13&0.21&\textcolor{blue}{0.71}&0.60&\textcolor{red}{0.74}&0.62&0.23&0.58&0.30&\textcolor{green}{0.66}\\
        \hline
        Singer1&\textcolor{green}{0.82}&0.57&0.25&0.75&0.36&0.79&0.80&\textcolor{red}{0.87}&0.45&0.19&0.36&\textcolor{blue}{0.83}\\ \hline
        Walking2&\textcolor{red}{0.82}&0.74&0.28&0.42&0.47&0.76&0.35&\textcolor{green}{0.78}&0.37&0.34&0.41&\textcolor{blue}{0.79}\\ \hline\hline
        Avg&0.52&0.37&0.35&0.41&0.40&0.61&0.44&0.51&0.39&0.40&0.32&\textcolor{red}{0.74}\\ \hline
       \multicolumn{9}{l}{\scriptsize Note: The best three results are highlighted in red, blue and green.}
    \end{tabular}
\end{table*}

\begin{table*}[thpb]
\centering
\small
\caption{Average tracking error}\label{cnterr}

\begin{tabular}{c c c c c c c c c c c c c c c c c} % centered columns (4 columns)
    \hline\hline
       & $L_1$APG & IVT & Frag & TLD & CSK & L2-RLS & ASLA& SCM& CXT&LOT &DFT&SHT \\ \hline\hline
       Bird2&80.4&28.4&28.1&221.6&\textcolor{blue}{17.9}&29.7&\textcolor{green}{19.6}&29.4&43.5&109.0&46.4&\textcolor{red}{10.0}\\ \hline
       Boy&17.8&91.8&33.9&4.0&20.1&\textcolor{green}{2.9}&89.6&53.1&\textcolor{red}{2.0}&64.7&106.3&\textcolor{blue}{2.7}\\
       \hline
       Caviar&21.7&64.1&27.2&43.6&71.4&\textcolor{green}{2.7}&62.1&\textcolor{blue}{2.5}&73.6&43.1&105.1&\textcolor{red}{2.4}\\ \hline
        Dog&11.3&93.7&12.1&51.5&6.9&11.6&9.5&\textcolor{green}{6.1}&\textcolor{blue}{5.9}&10.1&15.8&\textcolor{red}{5.6}\\ \hline
        DragB&129.7&92.8&46.3&213.1&87.9&\textcolor{green}{32.2}&65.7&62.8&89.1&\textcolor{blue}{26.3}&75.5&\textcolor{red}{8.6}\\ \hline
        Girl& 22.1&20.8&20.1&10.9&19.3&19.4&\textcolor{red}{4.2}&63.9&\textcolor{green}{6.2}&22.9&23.9&\textcolor{blue}{4.4}\\ \hline
        Jogging&\textcolor{red}{2.9}&89.3&27.5&\textcolor{green}{7.7}&164.7&37.0&175.0&141.5&120.4&14.3&33.4&\textcolor{blue}{3.5}\\
        \hline
        MounB&24.8&\textcolor{blue}{7.3}&208.7&208.5&\textcolor{red}{6.5}&18.4&\textcolor{green}{7.8}&11.7&178.7&24.9&155.0&10.9\\
        \hline
        Singer1&\textcolor{blue}{3.2}&11.7&77.0&8.3&14.0&4.1&\textcolor{blue}{3.2}&\textcolor{red}{2.9}&11.9&141.1&18.7&\textcolor{green}{3.5}\\ \hline
        Walking2&3.3&\textcolor{green}{3.1}&57.3&24.0&17.9&\textcolor{blue}{2.6}&38.1&\textcolor{red}{1.9}&33.0&64.6&29.0&\textcolor{blue}{2.6}\\ \hline\hline
        Avg&31.7&50.3&53.8&79.3&42.7&16.1&47.5&37.6&56.4&52.1&60.9&\textcolor{red}{5.5}\\ \hline
       \multicolumn{9}{l}{\scriptsize Note: The best three results are highlighted in red, blue and green.}
    \end{tabular}
\end{table*}

\begin{figure}[thpb]
  % Requires \usepackage{graphicx}
  \centering
\subfigure {\includegraphics[width=1\hsize]{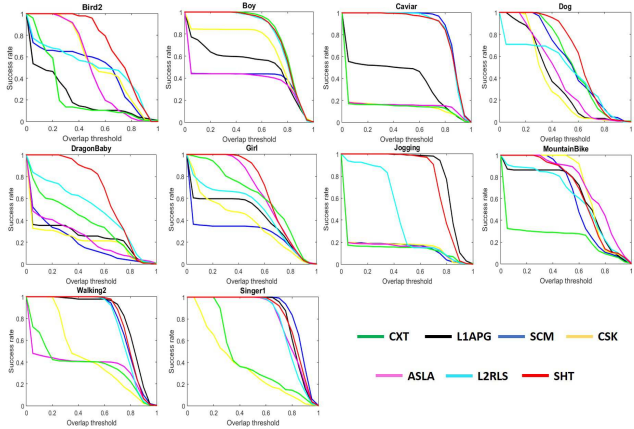} }
  \caption{Success plots of OPE. Seven trackers are tested on the ten datasets} \label{OPEval}
\end{figure}
For the purpose of thorough evaluation, the proposed SHT tracker is compared with other eleven algorithms in terms of overlap rate and average center error as presented in Table \ref{overlap} and \ref{cnterr} where the top three results are highlighted in red, blue and green fonts. The template size of L1APG is shifted from the default $12\times15$ to $32\times32$ for better performance at the cost of longer computational time. It can be observed that the SHT performs superiorly over other state-of-the-art algorithms. Specifically, compared with L2-RLS tracker, an obvious performance improvement can be achieved. Although, it already performs favorably on the datasets such as \emph{Boy} and \emph{Caviar} as stated in their paper, the proposed tracker further enhances the accuracy. The same situation holds for some other video sequences that can be successfully tracked by L2 algorithm. The performance is believed to be further improved if other embedded tracker with better robustness and accuracy is employed to replace L2. But the balance should also be considered between accuracy and speed. Moreover, different from saliency guided global search and superpixel HSV matching which need three channels (RGB) of the image, the linear refinement can be applied to all the particle filter based trackers because it only relies on greyscale image.\\
\indent We also run the One-Pass Evaluation (OPE) on the success rate $S(\tau_s)$ for seven trackers for comparison. Success rate can be defined as \cite{liu2014spatial}
\begin{eqnarray}
S(\tau_s)=\displaystyle\frac{\text{NUM}\{t|P_{or}(t)\geq \tau_s\}}{\text{length of the sequence}}
\end{eqnarray}
where $\tau_s$ is a threshold ranging from 0 to 1 with the interval of 0.1. $\text{NUM}\{\bullet\}$ returns the number of element in set $\{\bullet\}$. $P_{or}(t)$ denotes the overlap rate of \emph{t}th frame. The success rate curves are plotted in Figure \ref{OPEval}. The results reveal that SHT algorithm can always guarantee a remarkable success rate with different threshold values.
\section{Conclusion}
In this paper, a saliency guided robust visual tracking algorithm has been proposed with hierarchical structure. In global search, nineteen feature maps are combined with pre-trained weights to construct the saliency map. Then the possible locations are determined by searching the connected regions on this map. A novel integration mechanism is designed to filter out the false targets and then pass the estimated result to local search. A superpixel based HSV histogram matching scheme is incorporated into an L2 tracker to involve the color distribution matching in the local search. Finally, a linear refinement approach is introduced to further rectify the result with selected promising candidates. A customized fast solver is designed for this approach. In experiment, by qualitatively and quantitatively comparing with eleven state-of-the-art algorithms on ten challenging video sequences, the superiority of the proposed tracker has been demonstrated.
% Can use something like this to put references on a page
% by themselves when using endfloat and the captionsoff option.
\ifCLASSOPTIONcaptionsoff
  \newpage
\fi

\scriptsize
\bibliographystyle{ieeetr}
\bibliography{ref}
\end{document}